\documentclass{article}

\usepackage{hyperref}       
\usepackage{url}            
\usepackage{booktabs}       
\usepackage{nicefrac}       
\usepackage{microtype}      

\usepackage{graphicx}
\usepackage{amsmath,amssymb,mathtools,bm,etoolbox}
\usepackage{caption}
\usepackage{subcaption}
\usepackage{xcolor}
\usepackage{wrapfig}

\usepackage{tikz}
\usetikzlibrary{fit,positioning}

\PassOptionsToPackage{numbers}{natbib}
\usepackage[final]{neurips_2018}

\usepackage{algorithm, algpseudocode}
\floatname{algorithm}{Algorithm}

\algtext*{EndFunction}

\newcommand{\remove}[1]{}

\newcommand{\bfv}[1]{{\bf #1}}

\newcommand{\deriv}[2]{\frac{\textup{d}#1}{\textup{d}#2}}
\newcommand{\inderiv}[2]{\textup{d}#1/\textup{d}#2}
\newcommand{\sderiv}[2]{\frac{\textup{d}^2#1}{\textup{d}#2^2}}

\newcommand{\pderivw}[2]{\frac{\partial#1}{\partial#2}}
\newcommand{\expect}[1]{\mathbb{E}\left[#1\right]}
\newcommand{\expectw}[2]{\mathbb{E}_{#1}\left[#2\right]}
\newcommand{\variance}[1]{\mathbb{V}\left[#1\right]}

\newcommand{\estmr}[1]{\hat{#1}}
\newcommand{\p}[1]{\textup{p}(#1)}

\title{Total stochastic gradient algorithms and applications in
  reinforcement learning}

\author{
  Paavo Parmas
  \\
  Neural Computation Unit\\
  Okinawa Institute of Science and Technology Graduate University\\
  Okinawa, Japan \\
  \texttt{paavo.parmas@oist.jp} \\
}

\begin{document}

\maketitle

\begin{abstract}
  Backpropagation and the chain rule of derivatives have been
  prominent; however, the total derivative rule
  has not enjoyed the same amount of attention. In this work we show
  how the total derivative rule leads to an intuitive visual
  framework for creating gradient estimators on graphical models. In
  particular, previous "policy gradient theorems" are easily
  derived. We derive new gradient estimators based on density
  estimation, as well as a likelihood ratio gradient, which "jumps" to
  an intermediate node, not directly to the objective function. We
  evaluate our methods on model-based policy gradient algorithms,
  achieve good performance, and present evidence towards
  demystifying the success of the popular PILCO algorithm
  \cite{deisenroth2011pilco}.
\end{abstract}

\section{Introduction}

A central problem in machine learning is estimating the gradient of
the expectation of a random variable with respect to the parameters of the
distribution
$\deriv{}{\zeta}\expectw{x\sim\p{x;\zeta}}{\phi(x)}$. Some examples
include: the gradient of the expected classification error of a model
over the data generating distribution, the gradient of the expected
evidence lower bound w.r.t. the variational parameters in variational
inference \cite{hoffman2013stochastic}, or the gradient of the
expected reward w.r.t. the policy parameters in reinforcement learning
\cite{sutton1998reinforcement}. Usually, such an estimator is needed
not just through a single computation, but through a computation
graph; a good overview of related problems is given by
\cite{schulman2015stocgraph}. Previously, Schulman et al. provided a
method to obtain gradient estimators on stochastic computation graphs
by differentiating a surrogate loss
\cite{schulman2015stocgraph}. While the work provided an elegant
method to obtain gradient estimators using automatic differentiation,
the resulting {\it stochastic computation graph} framework has formal rules,
which uniquely define one specific type of estimator, and it is not
suitable for describing general gradient estimation techniques. For
example, determinstic policy gradients \cite{silver2014deterministic} or
total propagation \cite{pipps}
are not covered by the framework. In contrast, in probabilistic inference,
the successful probabilistic
graphical model framework \cite{pearl2014probabilistic}
only describes the structure of a model, while there are many different
choices of algorithms to perform inference. We aim for a similar
framework for gradient computation, which we call {\it probabilistic
  computation graphs}. Our framework uses the total derivative rule
$\deriv{f}{a} = \pderivw{f}{a} + \pderivw{f}{b}\deriv{b}{a}$ to
decompose the gradient into a sum of partial derivatives along
different computational paths, while leaving open the choice of estimator
for the partial derivatives.
We begin by introducing typical gradient
estimators in the literature, then explain our new
theorem, novel estimators using a non-standard decomposition
of the total derivative, and experimental results.

\paragraph{Nomenclature} All variables will be considered as
column vectors, and gradients are represented as matrices where each
row corresponds to one output variable, and each column corresponds to
one input variable---this allows applying the chain rule by simple
matrix multiplication, i.e.
$\deriv{f(\bfv{x})}{\bfv{y}} = \pderivw{f}{\bfv{x}}\pderivw{\bfv{x}}{\bfv{y}}$.
Matrices are vectorised with the $\textup{vec}(*)$ operator, i.e.
$\deriv{\Sigma}{\bfv{x}}$ means $\deriv{\textup{vec}(\Sigma)}{\bfv{x}}$.

\section{Background: Gradients of expectations}
\label{background}

\subsection{Pathwise derivative estimators}
This type of estimator relies on gradients of $\phi$ w.r.t. $\bfv{x}$,
e.g. the Gaussian gradient identities:
$\deriv{}{\mu}\expectw{\bfv{x}\sim\mathcal{N}(\mu,\Sigma)}
{\phi(\bfv{x})} = 
\expectw{\bfv{x}\sim
  \mathcal{N}(\mu,\Sigma)}{\deriv{\phi(\bfv{x})}{\bfv{x}}}$
and
$\deriv{}{\Sigma}\expectw{\bfv{x}\sim\mathcal{N}(\mu,\Sigma)} {\phi(\bfv{x})} =
\frac{1}{2}\expectw{\bfv{x}\sim
  \mathcal{N}(\mu,\Sigma)}{\sderiv{\phi(\bfv{x})}{\bfv{x}}}$, cited in
\cite{rezende2014stochasticBP}. The most prominent type of
pathwise derivative estimator are reparameterization (RP) gradients.
We focus our discussion on RP gradients, but we mentioned the
Gaussian identities to emphasize that RP gradients are not the only
possible pathwise estimators, e.g. the derivative w.r.t. $\Sigma$
given above does not correspond to an RP gradient. See
\cite{rezende2014stochasticBP} for an overview of various options.

\paragraph{RP gradient for a univariate Gaussian} To sample from
$\mathcal{N}(\mu, \sigma^2)$, sample from a standard normal
$\epsilon \sim \mathcal{N}(0,1)$, then transform this:
$x = \mu + \sigma\epsilon$. The gradients are $\inderiv{x}{\mu} = 1$
and $\inderiv{x}{\sigma} = \epsilon$. The gradient can then be
estimated by sampling:
$\deriv{}{\zeta}\expect{\phi(x)} = \expect{\deriv{\phi(x)}{x}
  \deriv{x}{\zeta}}$. For multivariate Gaussians, one can use the
Cholesky factor $L$ of $\Sigma= LL^T$ instead of $\sigma$. To
differentiate the Cholesky decomposition see
\cite{murray2016choldiv}. See \cite{rezende2014stochasticBP} for other
distributions. For a general distribution $\p{\bfv{x};\zeta}$, the RP
gradient defines a sampling procedure $\epsilon \sim \p{\epsilon}$ and a
transformation $\bfv{x} = f(\zeta,\epsilon)$, which allows moving the
derivative inside the expectation
$\deriv{}{\zeta}\expectw{\bfv{x}\sim\p{\bfv{x};\zeta}} {\phi(\bfv{x})}
= \expectw{\epsilon\sim\p{\epsilon}}{\deriv{\phi}{f}\deriv{f}{\zeta}}$.  The RP
gradient allows backpropagating the gradient through sampling
operations in a graph. It computes \emph{partial derivatives} through
a specific operation.

\subsection{Jump gradient estimators}

We introduce the categorization of {\it jump gradient estimators}. Unlike
pathwise derivatives, which compute local partial derivatives and
apply the chain rule through numerous computations, jump gradient
estimators can estimate the \emph{total derivative} directly using only
local computations---hence the naming: the gradient estimator jumps
over multiple nodes in a graph without having to differentiate
the nodes inbetween (this will become clearer in later sections in
the paper).

\paragraph{Likelihood ratio estimators (LR)}

Any function $f(\bfv{x})$ can be stochastically integrated by sampling
from an arbitrary distribution $\textup{q}(\bfv{x})$:
$\int f(\bfv{x})\textup{d}\bfv{x} = \int \textup{q}(\bfv{x})
\frac{f(\bfv{x})}{\textup{q}(\bfv{x})}\textup{d}\bfv{x} =
\expectw{\bfv{x}\sim \textup{q}}{f(\bfv{x})/\textup{q}(\bfv{x})}$.
The gradient of an expectation can be written as
$\int \phi(\bfv{x})
\deriv{\p{\bfv{x};\zeta}}{\zeta}\textup{d}\bfv{x}$. By picking
$\textup{q}(\bfv{x}) = \p{\bfv{x}}$, and stochastically integrating,
one obtains the LR gradient estimator:
$\expect{\frac{\inderiv{\p{\bfv{x};\zeta}}{\zeta}}
  {\p{\bfv{x};\zeta}}\phi(\bfv{x})}$. One \emph{must} subtract a
baseline from the $\phi(\bfv{x})$ values for this estimator to have
acceptable variance:
$\expect{\frac{\inderiv{\p{\bfv{x};\zeta}}{\zeta}}
  {\p{\bfv{x};\zeta}}(\phi(\bfv{x})-b)}$. In practice using
$b = \expect{\phi}$ is a reasonable choice. If $b$ does not depend on
the samples, then this leads to an unbiased gradient estimator.
Leave-one-out baseline estimates can be performed to achieve an
unbiased gradient estimator \cite{mnih2016looLR}. Other control
variate techniques also exist, and this is an active area of research
\cite{greensmith2004controlvariates}.

In our recent work \cite{pipps}, we introduced the batch
importance weighted LR estimator (BIW-LR)
and baselines:
{\bf BIW-LR:} $
\sum_{i=1}^P\sum_{j=1}^P
\left(\frac{\text{d}{\p{\bfv{x}_{j};\zeta_i(\theta)}}/\text{d}{\theta}}
  {\sum_{k=1}^P\p{\bfv{x}_{j};\zeta_k}}(\phi(\bfv{x}_{j}) - b_{i})\right)/P
$, where we use a mixture distribution $\textup{q} = \sum_{i}^P\p{\bfv{x};\zeta_i}/P$,
and each $\zeta_i$ depends on another set of parameters
$\theta$ (in our case the policy parameters), {\bf BIW-Baseline:}
$b_{i} = \left(\sum_{j\neq i}^P
  c_{j,i}\phi(\bfv{x}_{j})\right)/\sum_{j\neq i}^Pc_{j,i}$, where the importance
weights are
$c_{j,i} = \p{\bfv{x}_{j};\zeta_{i}}/
\sum_{k=1}^P\p{\bfv{x}_{j};\zeta_{k}}$.

\paragraph{Value function based estimators} 
Instead of using $\phi(\bfv{x})$ directly, one can learn an
approximator $\estmr{\phi}(\bfv{x})$. The approximator will often
require less computational time to evaluate, and could be used for
estimating the derivatives. Both LR gradients and pathwise derivatives
could be used with evaluations from the approximator. Moreover,
it is not necessary to evaluate just one $\bfv{x}$ point of
the estimator, but one could either use a larger number of samples, or
try to directly compute the expectation---this leads to a
Rao-Blackwellized estimator, which is known to have lower
variance. Such estimators have been considered for example in RL in
expected sarsa \cite{van2009theoretical, sutton1998reinforcement} as
well as in the stochastic variational inference literature
\cite{aueb2015local, tokui2017evaluating}, and also in policy
gradients \cite{ciosek2017expected, asadi2017mean}.

\section{Total stochastic gradient theorem}
\label{tottheorem}

Sec.~\ref{background} explained how to obtain estimators of the
expectation through a single computation, while here we explain how to
decompose the gradient of a complicated graph of computations into
smaller sections, which can be readily estimated using the methods in
Sec.~\ref{background}.  In our framework, we work with the gradient of
the marginal distribution. This more
general problem directly gives one the gradient of the expectation as
well, as the expectation is just a function of the marginal
distribution.

\subsection{Explanation of framework}
\newtheorem{compdef}{Definition}

We define {\it probabilistic computation graphs} (PCG). The
definition is exactly equivalent to the definition of a standard
directed graphical model, but it highlights our methods better, and
emphasizes our interest in computing gradients, rather than performing
inference.  The main difference is the explicit inclusion of the {\it
  distribution parameters} $\zeta$, e.g. for a Gaussian, the mean
$\mu$ and covariance $\Sigma$.

\begin{compdef}[Probabilistic computation graph (PCG)]
  An acyclic graph with nodes/vertices
  $V$ and edges $E$, which satisfy the following
  properties:
  \begin{enumerate}
  \item Each node $i\in V$ corresponds to a collection of random variables
    with marginal joint probability density $\textup{p}(\bfv{x}_i;\zeta_i)$,
    where $\zeta_i$ are the possibly infinite parameters of the
    distribution. Note that the parameterization is not unique,
    and any parameterization is acceptable.
  \item The probability density at each node is conditionally
    dependent on the parent nodes: $\textup{p}(\bfv{x}_i|\bfv{Pa}_i)$
    where $\bfv{Pa}_i$ are the random variables at the direct parents
    of node $i$.
  \item The joint probability density satisfies:
    $\textup{p}(\bfv{x}_1,...,\bfv{x}_n) =
    \prod_{i=1}^n\textup{p}(\bfv{x}_i|\bfv{Pa}_i)$
  \item Each $\zeta_i$ is a function of its parents:
    $\zeta_i = f(\bfv{Pz}_i)$ where $\bfv{Pz}_i$ are the distribution
    parameters at the parents of node i. In particular:
    $\textup{p}(\bfv{x}_i; \zeta_i) = \int
    \textup{p}(\bfv{x}_i|\bfv{Pa}_i)\textup{p}(\bfv{Pa}_i;\bfv{Pz}_i)\textup{d}
    \bfv{Pa}_i$
  \end{enumerate}
\end{compdef}

We emphasize that there is nothing stochastic in our formulation. Each
computation is determinstic, although they may be analytically
intractable. We also emphasize that this definition does not exclude
deterministic nodes, i.e. the distribution at a node may be a
Dirac delta distribution (a point mass). Later we will use this
formulation to derive stochastic estimates of the gradients.

\subsection{Derivation of theorem}
\label{theoderiv}
We are interested in computing the total derivative of the
distribution parameters at one node $\zeta_i$ w.r.t. the parameters at
another node $\text{d}\zeta_i/\text{d}\zeta_j$, e.g. nodes $i$ and $j$
could correspond to $\phi$ and $\bfv{x}$ in Sec.~\ref{background}
respectively. By the total derivative rule:
$\deriv{\zeta_i}{\zeta_j} = \sum_{\zeta_m\in
  \bfv{Pz}_i}\pderivw{\zeta_i}{\zeta_m}\deriv{\zeta_m}{\zeta_j}$.
Iterating this equation on the $\text{d}\zeta_m/\text{d}\zeta_j$ terms
leads to a sum over paths from node $j$ to node $i$:

\begin{equation}
  \label{eq:tot1}
  \deriv{\zeta_i}{\zeta_j} = \sum_{Paths(j\rightarrow i)}~~~
  \prod_{Edges (k,l)\in Path}\pderivw{\zeta_l}{\zeta_k}
\end{equation}

This equation holds for any deterministic computation graph, and is
also well known in e.g. the OJA
community \cite{naumann2008optimal}.
This equation trivially leads to our {\it total stochastic gradient
  theorem}, which states that the sum over paths from A to B can be
written as a sum over paths from A to intermediate nodes and from
the intermediate nodes to B. Fig.~\ref{totpaths} provides examples
of the paths in Eq.~\ref{eq:tot2} below.

\newtheorem{totstoc}{Theorem}

\begin{totstoc}[Total stochastic gradient theorem]
  Let $i$ and $j$ be distinct nodes in a probabilistic computation
  graph, and let $IN$ be any set of intermediate nodes, which block
  the paths from $j$ to $i$, i.e. $IN$ is such that there does not
  exist a path from $j$ to $i$, which does not pass through a node in
  $IN$. We denote $\{a\rightarrow b\}$ is the set of paths from $a$ to
  $b$, and $\{a\rightarrow b\}/c$ is the set of paths from $a$ to $b$,
  where no node along the path except for $b$ is allowed to be in set
  c. Then the total derivative $\textup{d}\zeta_i/\textup{d}\zeta_j$
  can be written with the equation below:
  
  \begin{equation}
    \label{eq:tot2}
    \deriv{\zeta_i}{\zeta_j} =  \sum_{m \in IN}
    \left(
      {\color{red}\left(\color{black} \sum_{s\in\{m\rightarrow i\}}~~~
        \prod_{(k,l)\in s} \pderivw{\zeta_l}{\zeta_k}\color{red}\right)}
      {\color{blue}\left(\color{black}\sum_{r\in\{j\rightarrow m\}/IN}~~~
        \prod_{(p,t)\in r}\pderivw{\zeta_t}{\zeta_p}\color{blue}\right)}
    \right)
  \end{equation}
  
\end{totstoc}

Equations~\ref{eq:tot1} and \ref{eq:tot2} can be combined to give:

\begin{equation}
    \label{eq:tot3}
    \deriv{\zeta_i}{\zeta_j} =  \sum_{m \in IN}
    \left(
      \left( \deriv{\zeta_i}{\zeta_m}
      \right)
      \left(\sum_{r\in\{j\rightarrow m\}/IN}~~~
        \prod_{(p,t)\in r}\pderivw{\zeta_t}{\zeta_p}\right)
    \right)
\end{equation}

Note that an analogous theorem could be derived by swapping
$r\in\{j\rightarrow m\}/IN$ and $s\in\{m\rightarrow i\}$ with
$r\in\{j\rightarrow m\}$ and $s\in\{m\rightarrow i\}/IN$ respectively. This
leads to the equation below:

\begin{equation}
    \label{eq:tot4}
    \deriv{\zeta_i}{\zeta_j} =  \sum_{m \in IN}
    \left(
      \left(\sum_{r\in\{m\rightarrow i\}/IN}~~~
        \prod_{(p,t)\in r}\pderivw{\zeta_t}{\zeta_p}\right)
      \left( \deriv{\zeta_m}{\zeta_j}      \right)
    \right)
\end{equation}

We will refer to Equations~\ref{eq:tot3} and \ref{eq:tot4} as the
second and first half {\it total gradient equations} respectively.

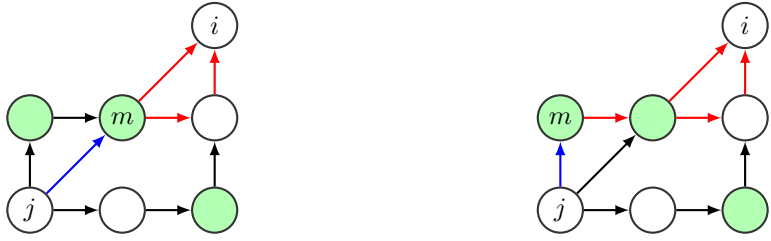
\begin{figure}
  \begin{subfigure}{.5\columnwidth}
    \centering
\begin{tikzpicture}
\tikzstyle{main}=[circle, minimum size = 6mm, thick, draw =black!80, node distance = 6mm]
\tikzstyle{connect}=[-latex, thick]
\tikzstyle{box}=[rectangle, draw=black!100]
  \node[main, fill = white!100] (j) [label=center:$j$] { };
  \node[main, fill = white!100] (r1) [right=of j] { };
  \node[main, fill = green!30] (r2) [right=of r1] {};
  \node[main, fill = green!30] (r21) [above=of j] { };
  \node[main, fill = green!30] (r22) [right=of r21, label=center:$m$] { };
  \node[main, fill = white!100] (r23) [right=of r22] { };
  \node[main, fill = white!100] (i) [above=of r23, label=center:$i$] { };
  \path (j) edge [connect] (r1)
        (r1) edge [connect] (r2)
		(r2) edge [connect] (r23)
		(j) edge [connect] (r21)
                (r23) edge [connect, color = red!100] (i)
                (r21) edge [connect] (r22)
                (r22) edge [connect, color = red!100] (r23)
                (j) edge [connect, color = blue!100] (r22)
                (r22) edge [connect, color = red!100] (i);
\end{tikzpicture}
\caption{$\{j\rightarrow m\}$ paths may not pass through
  green nodes.}
\end{subfigure}
\begin{subfigure}{.5\columnwidth}
  \centering
\begin{tikzpicture}
\tikzstyle{main}=[circle, minimum size = 6mm, thick, draw =black!80, node distance = 6mm]
\tikzstyle{connect}=[-latex, thick]
\tikzstyle{box}=[rectangle, draw=black!100]
  \node[main, fill = white!100] (j) [label=center:$j$] { };
  \node[main, fill = white!100] (r1) [right=of j] { };
  \node[main, fill = green!30] (r2) [right=of r1] {};
  \node[main, fill = green!30] (r21) [above=of j, label=center:$m$] { };
  \node[main, fill = green!30] (r22) [right=of r21] { };
  \node[main, fill = white!100] (r23) [right=of r22] { };
  \node[main, fill = white!100] (i) [above=of r23, label=center:$i$] { };
  \path (j) edge [connect] (r1)
        (r1) edge [connect] (r2)
		(r2) edge [connect] (r23)
		(j) edge [connect, color = blue!100] (r21)
                (r23) edge [connect, color = red!100] (i)
                (r21) edge [connect, color = red!100] (r22)
                (r22) edge [connect, color = red!100] (r23)
                (j) edge [connect] (r22)
                (r22) edge [connect, color = red!100] (i);
\end{tikzpicture}
\caption{$\{m\rightarrow i\}$ paths may pass through green nodes.}
\end{subfigure}
\caption{Example paths in Equation~\ref{eq:tot2}. The green nodes
correspond to the intermediate nodes $IN$.}
\label{totpaths}
\end{figure}

\subsection{Gradient estimation on a graph}
\label{seq:ancgrad}
Here we clarify one method how the partial derivatives through the
nodes $m \in IN$ in the previous section can be estimated. We use the following properties of
the estimators in Sec.~\ref{background}:
\begin{itemize}
\item \emph{Pathwise derivative estimators} compute partial derivatives through
  a single edge, 
  e.g. $\pderivw{\zeta_m}{\zeta_j}$
\item \emph{Jump gradient estimators} sum the gradients across all computational
  paths between two nodes and directly compute total derivatives, e.g. $\deriv{\zeta_i}{\zeta_m}$
\end{itemize}
The task is to estimate the derivative of the expectation at a distal
node $i$ w.r.t. the parameters at an earlier node $j$:
$\deriv{}{\zeta_j} \expectw{\bfv{x}_i \sim \p{\bfv{x}_i;\zeta_i}}
{\bfv{x}_i}$, through an intermediate node $m$. Note that
$\expect{\bfv{x}_i}$ can be picked as one of the distribution
parameters in $\zeta_i$.  The true $\zeta$ are intractable, so we
perform an ancestral sampling based estimate $\estmr{\zeta}$, i.e. we
sample sequentially from each $\textup{p}(\bfv{x}_*|\textup{Pa}_*)$ to
get a sample through the whole graph, then $\estmr{\zeta}_*$ will
simply be the parameters of $\textup{p}(\bfv{x}_*|\textup{Pa}_*)$.  We
refer to one such sample as a {\it particle}. We use a batch of $P$
such particles $\estmr{\zeta}_* = \{\estmr{\zeta}_{*,c}\}_c^P$ to
obtain a mixture distribution as an approximation to the true
distribution.  Such a sampling procedure has the properties
$\p{\bfv{x};\zeta} = \int \p{\bfv{x};\estmr{\zeta}}\p{\estmr{\zeta}}
\textup{d}\estmr{\zeta}$ and
$\expectw{\bfv{x}_i \sim \p{\bfv{x}_i;\zeta_i}}{\bfv{x}_i} =
\expectw{\estmr{\zeta}_i\sim \p{\estmr{\zeta}_i;\zeta_j}}
{\expectw{\bfv{x}_i \sim \p{\bfv{x}_i;\estmr{\zeta}_i}}{\bfv{x}_i}}$.
For simplicity in the explanation, we further assume that the sampling
is reparameterizable, i.e.
$\p{\estmr{\zeta}_m;\zeta_j} = \int f(\estmr{\zeta}_m;\zeta_j,
\epsilon_m)\p{\epsilon_m}\textup{d}\epsilon_m$.  We can write
$\deriv{}{\zeta_j} \expectw{\estmr{\zeta}_i\sim
  \p{\estmr{\zeta}_i;\zeta_j}} {\expectw{\bfv{x}_i \sim
    \p{\bfv{x}_i;\estmr{\zeta}_i}}{\bfv{x}_i}} =
\expectw{\epsilon_m\sim \p{\epsilon_m}}
{\pderivw{\estmr{\zeta}_m}{\zeta_j}\deriv{}{\estmr{\zeta}_m}
  \expectw{\bfv{x}_i \sim
    \p{\bfv{x}_i;\estmr{\zeta}_i}}{\bfv{x}_i}}$. The term
$\pderivw{\estmr{\zeta}_m}{\zeta_j}$ will be estimated with a pathwise
derivative estimator. The remaining term
$\deriv{}{\estmr{\zeta}_m} \expectw{\bfv{x}_i \sim
  \p{\bfv{x}_i;\estmr{\zeta}_i}}{\bfv{x}_i}$ will be estimated with
any other estimator, e.g. a jump estimator could be used.

We summarize the procedure for creating gradient estimators from $j$
to $i$ on the whole graph:
\begin{enumerate}
\item Choose a set of intermediate nodes $IN$, which block the paths from $j$
  to $i$.
\item Construct pathwise derivative estimators from $j$ to the intermediate
  nodes $IN$.
\item Construct total derivative estimators from $IN$ to $i$, and apply
  Eq.~\ref{eq:tot3} to combine the gradients.
\end{enumerate}

\section{Relationship to policy gradient theorems}
\label{reltopol}
\begin{figure}
\begin{subfigure}{.5\columnwidth}
\centering
\begin{tikzpicture}
\tikzstyle{main}=[circle, minimum size = 6mm, thick, draw =black!80, node distance = 4mm]
\tikzstyle{connect}=[-latex, thick]
\tikzstyle{box}=[rectangle, draw=black!100]
  \node[main, fill = white!100] (x0) [label=center:$\bfv{x}_0$] { };
  \node[main, fill = white!100] (x1) [right=of x0,label=center:$\bfv{x}_1$] {};
  \node[main, fill = green!30] (u0) [below=of x0,label=center:$\bfv{u}_0$] {};
  \node[main, fill = green!30] (u1) [below=of x1, label=center:$\bfv{u}_1$] {};
  \node[main, fill = white!100] (x2) [right=of x1, label=center:$\bfv{x}_2$] {};
  \node[main, fill = green!30] (u2) [below=of x2, label=center:$\bfv{u}_2$] {};
  \node[main, fill = white!100] (x3) [right=of x2, label=center:$\bfv{x}_3$] {};
  \node[main, fill = white!100] (c1) [above=of x1, label=center:$c_1$] {};
  \node[main, fill = white!100] (c2) [above=of x2, label=center:$c_2$] {};
  \node[main, fill = white!100] (c3) [above=of x3, label=center:$c_3$] {};
  \node[main, fill = white!100] (G) [above=of c2, label=center:$G$] {};
  \node[main, fill = white!100] (theta) [below=of u1, label=center:$\theta$] {};  
  \path (x0) edge [connect] (x1)
        (x0) edge [connect] (u0)
		(x1) edge [connect] (u1)
		(u0) edge [connect] (x1)
                (x1) edge [connect] (x2)
                (x2) edge [connect, color=red!100] (x3)
                (x2) edge [connect, color=red!100] (u2)
                (u1) edge [connect, color=red!100] (x2)
                (u2) edge [connect, color=red!100] (x3)                
                (x1) edge [connect] (c1)
                (x2) edge [connect, color=red!100] (c2)
                (x3) edge [connect, color=red!100] (c3)
                (c1) edge [connect] (G)
                (c2) edge [connect, color=red!100] (G)
                (c3) edge [connect, color=red!100] (G)
                (theta) edge [connect] (u0)
                (theta) edge [connect, color=blue!100] (u1)
                (theta) edge [connect] (u2);
\end{tikzpicture}
\caption{Classical model-free policy gradient}
\label{modelfreepol}
\end{subfigure}
\begin{subfigure}{.5\columnwidth}
  \centering
\begin{tikzpicture}
\tikzstyle{main}=[circle, minimum size = 6mm, thick, draw =black!80, node distance = 4mm]
\tikzstyle{connect}=[-latex, thick]
\tikzstyle{box}=[rectangle, draw=black!100]
  \node[main, fill = white!100] (x0) [label=center:$\bfv{x}_0$] { };
  \node[main, fill = green!30] (x1) [right=of x0,label=center:$\bfv{x}_1$] {};
  \node[main, fill = white!100] (u0) [below=of x0,label=center:$\bfv{u}_0$] {};
  \node[main, fill = white!100] (u1) [below=of x1, label=center:$\bfv{u}_1$] {};
  \node[main, fill = green!30] (x2) [right=of x1, label=center:$\bfv{x}_2$] {};
  \node[main, fill = white!100] (u2) [below=of x2, label=center:$\bfv{u}_2$] {};
  \node[main, fill = green!30] (x3) [right=of x2, label=center:$\bfv{x}_3$] {};
  \node[main, fill = white!100] (c1) [above=of x1, label=center:$c_1$] {};
  \node[main, fill = white!100] (c2) [above=of x2, label=center:$c_2$] {};
  \node[main, fill = white!100] (c3) [above=of x3, label=center:$c_3$] {};
  \node[main, fill = white!100] (G) [above=of c2, label=center:$G$] {};
  \node[main, fill = white!100] (theta) [below=of u1, label=center:$\theta$] {};  
  \path (x0) edge [connect] (x1)
        (x0) edge [connect] (u0)
		(x1) edge [connect] (u1)
		(u0) edge [connect] (x1)
                (x1) edge [connect] (x2)
                (x2) edge [connect, color=red!100] (x3)
                (x2) edge [connect, color=red!100] (u2)
                (u1) edge [connect, color=blue!100] (x2)
                (u2) edge [connect, color=red!100] (x3)                
                (x1) edge [connect] (c1)
                (x2) edge [connect, color=red!100] (c2)
                (x3) edge [connect, color=red!100] (c3)
                (c1) edge [connect] (G)
                (c2) edge [connect, color=red!100] (G)
                (c3) edge [connect, color=red!100] (G)
                (theta) edge [connect] (u0)
                (theta) edge [connect, color=blue!100] (u1)
                (theta) edge [connect] (u2);
\end{tikzpicture}
\caption{Model-based state-space LR gradient}
\label{modelbasedpol}
\end{subfigure}
\caption{Probabilistic computation graphs for model-based and
  model-free LR gradient estimation.}
\label{LRPCG}
\end{figure}
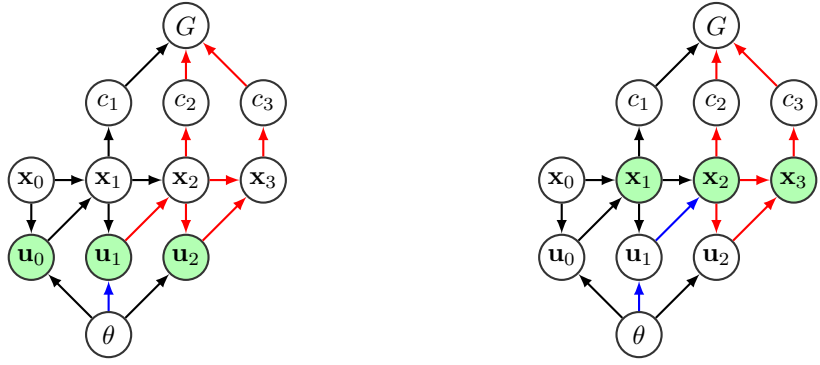

In typical model-free RL problems \cite{sutton1998reinforcement} an
agent performs actions $\bfv{u} \sim \pi(\bfv{u}_t|\bfv{x}_t;\theta)$ according to a
stochastic policy $\pi$, transitions through states $\bfv{x}_t$, and obtains
costs $c_t$ (or conversely rewards). The agent's goal is to find the
policy parameters $\theta$, which optimize the expected return
$G = \sum_{t=0}^H c_t$ for each episode. The corresponding probabilistic
computation graph is provided in Fig.~\ref{modelfreepol}.

In the literature, two "gradient theorems" are widely applied: the
policy gradient theorem \cite{sutton2000policy}, and the deterministic
policy gradient theorem \cite{silver2014deterministic}. These two are
equivalent in the limit of no noise
\cite{silver2014deterministic}.

\paragraph{Policy gradient theorem}
\begin{equation}
  \label{polgrad}
  \deriv{}{\theta}\expect{G} = \expect{\sum_{t=0}^{H-1}
    \deriv{\log \pi(\bfv{u}_t|\bfv{x}_t;\theta)}{\theta}
    \estmr{Q}_t(\bfv{u}_t,\bfv{x}_t)}
\end{equation}

\paragraph{Deterministic policy gradient theorem}
\begin{equation}
  \label{detpolgrad}
  \deriv{}{\theta}\expect{G} = \expect{\sum_{t=0}^{H-1}
    \deriv{\bfv{u}_t}{\theta}\deriv{\estmr{Q}_t(\bfv{u}_t,\bfv{x}_t)}
    {\bfv{u}_t}}
\end{equation}

$\estmr{Q}_t$ corresponds to an estimator of the remaining return
$\sum_{h=t}^{H-1}c_{h+1}$ from a particular state $\bfv{x}$ when
choosing action $\bfv{u}$. For Eq.~\ref{polgrad} any estimator is
acceptable, even a sample based estimate could be used. For
Eq.~\ref{detpolgrad}, $\estmr{Q}$ is usually a differentiable
surrogate model.  Fig.~\ref{modelfreepol} shows how these two theorems
correspond to the same probabilistic computation graph. The
intermediate nodes are the actions selected at each time step. The
difference lies in the choice of jump estimator to estimate the total
derivative following the intermediate nodes---the policy gradient
theorem uses an LR gradient, whereas the deterministic policy gradient
theorem uses a pathwise derivative to a surrogate model. We believe
that the derivation based on a PCG is more intuitive than previous
algebraic proofs \cite{sutton2000policy,silver2014deterministic}.

\section{Novel algorithms}
\label{novintro}
In Sec.~\ref{seq:ancgrad} we explained how a particle-based mixture
distribution is used for creating gradient estimators. In the
following sections, we instead take advantage of these particles to
estimate a different parameterization $\Gamma$, directly for the
marginal distribution. Although the algorithms have general
applicability, to make a concrete example, we explain them in
reference to model-based policy gradients using a differentiable model
considered in our previous work \cite{pipps}, for which the PCG is
given in Fig.~\ref{modelbasedpol}. Stochastic value gradients
\cite{heess2015learning}, for example, share the same PCG.

\subsection{Density estimation LR (DEL)}
\label{DEL}
Following the explanation in Sec.~\ref{novintro},
one could attempt to estimate the distribution parameters $\Gamma$
from a set of sampled particles, then apply the LR gradient using the
estimated distribution $\textup{q}(\bfv{x};\Gamma)$. In particular, we will
approximate the density as a Gaussian by estimating the mean
$\estmr{\mu} = \sum_i^P\bfv{x}_i/P$ and variance
$\estmr{\Sigma} = \sum_i^P(\bfv{x}_i-\estmr{\mu})^2/(P-1)$. Then, using
the standard LR trick, one can estimate the gradient
$\sum_i^P\deriv{\log \textup{q}(\bfv{x}_i)}{\theta}(G_i-b)$,
where $\textup{q}(\bfv{x})=\mathcal{N}(\estmr{\mu},\estmr{\Sigma})$.
To use this method, one must compute derivatives of
$\estmr{\mu}$ and $\estmr{\Sigma}$ w.r.t. the particles $\bfv{x}_i$, then
carry the gradient to the policy parameters using the chain rule while
differentiating through the model, 
which is straight-forward. We refer to our new method
as the DEL estimator. Importantly, note that while $\textup{q}(\bfv{x})$
is used for estimating the gradient, it is not in any way used for
modifying the trajectory sampling.\\
{\bf Advantages of DEL:} One can use LR gradients
even if no noise is injected into the computations. \\
{\bf Disadvantages of DEL:}
The estimator is biased, and density estimation can be difficult.

\subsection{Gaussian shaping gradient (GS)}
\label{GS}

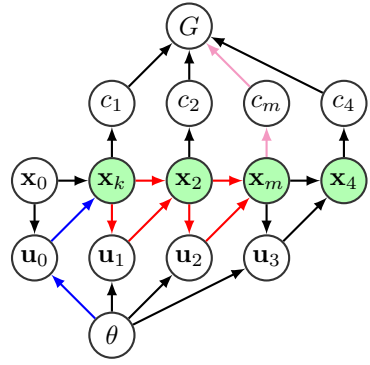
\begin{wrapfigure}[15]{r}{0.40\linewidth}
\centering
\begin{tikzpicture}
\tikzstyle{main}=[circle, minimum size = 6mm, thick, draw =black!80, node distance = 4mm]
\tikzstyle{connect}=[-latex, thick]
\tikzstyle{box}=[rectangle, draw=black!100]
  \node[main, fill = white!100] (x0) [label=center:$\bfv{x}_0$] { };
  \node[main, fill = green!30] (x1) [right=of x0,label=center:$\bfv{x}_k$] {};
  \node[main, fill = white!100] (u0) [below=of x0,label=center:$\bfv{u}_0$] {};
  \node[main, fill = white!100] (u1) [below=of x1, label=center:$\bfv{u}_1$] {};
  \node[main, fill = green!30] (x2) [right=of x1, label=center:$\bfv{x}_2$] {};
  \node[main, fill = white!100] (u2) [below=of x2, label=center:$\bfv{u}_2$] {};
  \node[main, fill = green!30] (x3) [right=of x2, label=center:$\bfv{x}_m$] {};
  \node[main, fill = white!100] (u3) [below=of x3, label=center:$\bfv{u}_3$] {};
  \node[main, fill = green!30] (x4) [right=of x3, label=center:$\bfv{x}_4$] {};
  \node[main, fill = white!100] (c1) [above=of x1, label=center:$c_1$] {};
  \node[main, fill = white!100] (c2) [above=of x2, label=center:$c_2$] {};
  \node[main, fill = white!100] (c3) [above=of x3, label=center:$c_m$] {};
  \node[main, fill = white!100] (c4) [above=of x4, label=center:$c_4$] {};
  \node[main, fill = white!100] (G) [above=of c2, label=center:$G$] {};
  \node[main, fill = white!100] (theta) [below=of u1, label=center:$\theta$] {};  
  \path (x0) edge [connect] (x1)
        (x0) edge [connect] (u0)
		(x1) edge [connect, color=red!100] (u1)
		(u0) edge [connect, color=blue!100] (x1)
                (x1) edge [connect, color=red!100] (x2)
                (x2) edge [connect, color=red!100] (x3)
                (x2) edge [connect, color=red!100] (u2)
                (x3) edge [connect] (u3)
                (x3) edge [connect] (x4)                
                (u1) edge [connect, color=red!100] (x2)
                (u2) edge [connect, color=red!100] (x3)                
                (u3) edge [connect] (x4)                
                (x1) edge [connect] (c1)
                (x2) edge [connect] (c2)
                (x3) edge [connect, color=magenta!50] (c3)
                (x4) edge [connect] (c4)
                (c1) edge [connect] (G)
                (c2) edge [connect] (G)
                (c3) edge [connect, color=magenta!50] (G)
                (c4) edge [connect] (G)
                (theta) edge [connect, color=blue!100] (u0)
                (theta) edge [connect] (u1)
                (theta) edge [connect] (u2)
                (theta) edge [connect] (u3);
              \end{tikzpicture}
              \caption{Computational paths in Gaussian shaping gradient}
              \label{GSpic}
\end{wrapfigure}

Until now, all RL methods have used the second half total gradient
equation (Eq.~\ref{eq:tot3}).
Might one create estimators that use
the first half equation (Eq.~\ref{eq:tot4})?
Fig.\ref{GSpic} gives an
example of how this might be done. We propose to estimate the density
at $\bfv{x}_m$ by fitting a Gaussian on the particles. Then
$\inderiv{\expect{c_m}}{\Gamma_m}$ (the pink edges) will be estimated
by sampling from this distribution (or by any other method of
integration). This leaves the question of how to estimate
$\inderiv{\Gamma_m}{\theta}$ (all paths from $\theta$ to $\bfv{x}_m$).
Using the RP method is straight-forward.  To use the LR method, we
first apply the second half total gradient equation on
$\inderiv{\Gamma_m}{\theta}$ to obtain terms
$\sum_{r\in\{\theta\rightarrow x_k\}/IN}\prod_{(p,t)\in
  r}\pderivw{\zeta_{t}}{\zeta_{p}}$ (blue edges) and
$\deriv{\Gamma_m}{\zeta_{x_k}}$ (red edges). In the scenarios we
consider, the first of these terms is a single path, and will be
estimated using RP.  The second term is more interesting, and we will
estimate this using an LR method.

As we are using a Gaussian approximation, the distribution parameters
$\Gamma_m$ are the mean and variance of $\bfv{x}_m$, which can be
estimated as $\mu_m = \expect{\bfv{x}_m}$ and
$\Sigma_m = \expect{\bfv{x}_m\bfv{x}_m^T} - \mu_m\mu_m^T$. We can
obtain LR gradient estimates of these terms
$\deriv{}{\zeta_{x_k}}\expect{\bfv{x}_m} = \expectw{\bfv{x}_k\sim
  \textup{p}(\bfv{x}_k;\zeta_{x_k})}
{\deriv{\log\textup{p}(\bfv{x_k};\zeta_{x_k})}{\zeta_{x_k}}(\bfv{x}_m-\bfv{b}_\mu)}$,
$\deriv{}{\zeta_{x_k}}\expect{\bfv{x}_m\bfv{x}_m^T} =
\expectw{\bfv{x}_k\sim \textup{p}(\bfv{x}_k;\zeta_{x_k})}
{\deriv{\log\textup{p}(\bfv{x_k};\zeta_{x_k})}{\zeta_{x_k}}(\bfv{x}_m\bfv{x}_m^T-
\bfv{b}_\Sigma)}$
and
$\deriv{}{\zeta_{x_k}}(\mu\mu^T) =
2\mu\deriv{}{\zeta_{x_k}}\expect{\bfv{x}_m^T}$.  In practice, we perform a
sampling based estimate $\estmr{\zeta}_{x_k}$, and one might be concerned
that the estimators are conditional on the sample $\estmr{\zeta}_{x_k}$,
but we are interested in unconditional estimates. We will explain that
the conditional estimate is equivalent. For the variance, note that
$\mu_m$ is an estimate of the unconditional mean, so the whole
estimate directly corresponds to an estimate of the unconditional
variance.  For the mean, apply the rule of iterated expectations:
$\expectw{\bfv{x}_k\sim\p{\bfv{x}_k;\zeta_{x_k}}}{\bfv{x}_m} =
\expectw{\estmr{\zeta}_{x_k}\sim \p{\estmr{\zeta}_{x_k}}}
{\expectw{\bfv{x}_k\sim\p{\bfv{x}_k;\estmr{\zeta}_{x_k}}}
  {\bfv{x}_m}}$ from which it is clear that the conditional gradient estimate
is an unbiased estimator for the gradient of the unconditional mean.

\paragraph{Efficient algorithm for accumulating gradients}
In Fig.~\ref{GSpic}, for
each $\bfv{x}_k$ node, we want to perform an LR jump to every
$\bfv{x}_m$ node after $k$ and compute a gradient with the Gaussian
approximation of the distribution at node $m$.  We will accumulate
across all nodes during a backwards pass in a backpropagation like
manner. Note that for each $k$ and each $m$, we can write the
gradient as
$\deriv{\expect{c_m}}{\Gamma_m}\deriv{\Gamma_m}{\zeta_{x_k}}
(\deriv{\zeta_{x_k}}{\bfv{u}_{k-1}}
\deriv{\bfv{u}_{k-1}}{\theta})$. The term
$\deriv{\expect{c_m}}{\Gamma_m}\deriv{\Gamma_m}{\zeta_{x_k}}$ is
estimated as
$\deriv{\expect{c_m}}{\Gamma_m}\bfv{z}_m
\deriv{\log\textup{p}(\bfv{x_k};\zeta_{x_k})}{\zeta_{x_k}}$, where
$\bfv{z}_m$ corresponds to a vector summarizing the
$\bfv{x}_m-\bfv{b}_\mu$, etc. terms above.  Note that
$\deriv{\expect{c_m}}{\Gamma_m}\bfv{z}_m$ is just a scalar quantity
$g_m$. We thus use an algorithm which accumulates a sum of all $g$
during a backwards pass, and sums over all $m$ nodes at each $k$
node. See Alg.~\ref{alg:total-prop} for a detailed explanation of how
it fits together with total propagation \cite{pipps}. The final
algorithm essentially just replaces the usual cost/reward with a
modified value, and such an approach would also be applicable in
model-free policy gradient algorithms using a stochastic policy and LR
gradients.

\paragraph{Two interpretations of GS} 1. We are making a Gaussian
approximation of the marginal distribution at a node. 2. We are
performing a type of reward shaping based on the distribution of the
particles. In particular we are essentially promoting the trajectory
distributions to stay unimodal, such that all of the particles
concentrate at one "island" of reward rather than splitting the
distribution between multiple regions of reward---this
may simplify optimization.

\begin{algorithm}[tb]
  \caption{Gaussian shaping gradient with total propagation}
   \label{alg:total-prop}
\begin{algorithmic}
  \State Gaussian shaping gradient for model-based
  policy search while combining both LR and RP variants using total
  propagation---an algorithm introduced in our previous work
  \cite{pipps}.  \State {\bfseries Forward pass:} Sample a set of
  particle trajectories.
  \State {\bfseries Backward pass:}\\
  \State {\bfseries Initialise:}
  $\deriv{G_{T+1}}{\zeta_{T+1}} = \bfv{0}$,
  $\deriv{J}{\theta} = \bfv{0}$, $G_{T+1} = 0$
  \Comment $\zeta$ are the distribution parameters, e.g.
  all of the $\mu$ and $\sigma$ for each particle

  \For{$t=T$
    {\bfseries to} $1$}
  \State $\mu_t = \expect{\bfv{x}_t}$; $\Sigma_t =
  \expect{\bfv{x}_t\bfv{x}_t^T} - \mu_t\mu_t^T$
  \Comment Estimate the marginal distribution as a Gaussian
  \State {\bfseries Compute:} $\deriv{\expect{c_t}}{\mu_t}$ and
  $\deriv{\expect{c_t}}{\Sigma_t}$, e.g. by sampling from this Gaussian,
  and using the RP gradient
  \For{{\bfseries each} particle $i$}
  \State $\bfv{m}_{i,t}$ = $\bfv{x}_{i,t}-\mu_t$; 
  $\bfv{v}_{i,t}$ = $\textup{vec}\left(\bfv{x}_{i,t}\bfv{x}_{i,t}^T
    - \expect{\bfv{x}_t\bfv{x}_t^T}\right)$;
  $\bfv{w}_{i,t} = \textup{vec}\left(\bfv{m}_{i,t}\mu_t^T\right)$
  \Comment $\textup{vec}(*)$ is a vectorization operator which stacks the elements
  in a matrix/tensor into a column vector
  \State 
  $g_{i,t} = \deriv{\expect{c_t}}{\mu_t}\bfv{m}_{i,t}
  + \deriv{\expect{c_t}}{\Sigma_t}(\bfv{v}_{i,t}-2\bfv{w}_{i,t})$
  \Comment $g$ is a scalar replacing the usual cost/reward
  \State
  $G_{i,t} = G_{i,t+1} + g_{i,t}$
  \Comment $G$ is the return (the cost of the remaining trajectory)
  \State 
  $\deriv{\expect{c_t}}{\bfv{x}_{i,t}} =
  \deriv{\expect{c_t}}{\mu_t}\deriv{\mu_t}{\bfv{x_{i,t}}} +
  \deriv{\expect{c_t}}{\Sigma_t}\deriv{\Sigma_t}{\bfv{x_{i,t}}}$
  \Comment Direct derivative of expected cost for the RP gradient
  \State 
  $\deriv{\bfv{\zeta}_{i,t+1}}{\bfv{x}_{i,t}} =
  \pderivw{\bfv{\zeta}_{i,t+1}}{\bfv{x}_{i,t}} +
  \deriv{\bfv{\zeta}_{i,t+1}}{\bfv{u}_{i,t}}\deriv{\bfv{u}_{i,t}}{\bfv{x}_{i,t}}$
  \State
  $\deriv{G_{i,t}^{RP}}{\zeta_{i,t}} =
  (\deriv{G_{i,t+1}}{\zeta_{i,t+1}}
  \deriv{\zeta_{i,t+1}}{\bfv{x}_{i,t}} +
  \deriv{\expect{c_t}}{\bfv{x}_{i,t}})\deriv{\bfv{x}_{i,t}}{\zeta_{i,t}}$
  \State
  $\deriv{G_{i,t}^{LR}}{\zeta_{i,t}} = G_{i,t}\deriv{\log
    \textup{p}(\bfv{x}_{i,t})}{\zeta_{i,t}}$
  \Comment In principle, one could further subtract a baseline from $G$
  \State 
  $\deriv{G_{i,t}^{RP}}{\theta} = \deriv{G_{i,t}^{RP}}
  {\zeta_{i,t}}\deriv{\zeta_{i,t}}{\bfv{u}_{i,t-1}}\deriv{\bfv{u}_{i,t-1}}{\theta}$
  \State 
  $\deriv{G_{i,t}^{LR}}{\theta} = \deriv{G_{i,t}^{LR}}
  {\zeta_{i,t}}\deriv{\zeta_{i,t}}{\bfv{u}_{i,t-1}}\deriv{\bfv{u}_{i,t-1}}{\theta}$
  \EndFor \State
  $\sigma^2_{RP} =
  \text{trace}(\variance{\deriv{G_{i,t}^{RP}}{\theta}})$;
  $\sigma^2_{LR} =
  \text{trace}(\variance{\deriv{G_{i,t}^{LR}}{\theta}})$
  \Comment The sample variance of the particles
  \State
  $k_{LR} = 1/\left(1+\frac{\sigma^2_{LR}}{\sigma^2_{RP}}\right)$
  \Comment Weight to combine LR and RP estimators
  \State
  $\deriv{J}{\theta} = \deriv{J}{\theta} + k_{LR}\frac{1}{P}\sum_i^P
  \deriv{G_{i,t}^{LR}}{\theta} + (1-k_{LR})\frac{1}{P}\sum_i^{P}
  \deriv{G_{i,t}^{RP}}{\theta}$
  \Comment Combine LR and RP in $\theta$ space
  \For{{\bfseries each} particle $i$}
  \State
  $\deriv{G_{i,t}}{\zeta_{i,t}} =
  k_{LR}\deriv{G_{i,t}^{LR}}{\zeta_{i,t}} +
  (1-k_{LR})\deriv{G_{i,t}^{RP}}{\zeta_{i,t}}$
  \Comment Combine LR and RP in state space
  \EndFor \EndFor
\end{algorithmic}
\end{algorithm}

\section{Experiments}

We performed model-based RL simulation experiments from the PILCO
papers \cite{deisenroth2011pilco, deisenroth2015gppilco}. We tested
the cart-pole swing-up and balancing problems to test our GS approach, as well as
combinations with total propagation \cite{pipps}. We also tested the DEL approach
on the simpler cart-pole balancing-only-problem to show the feasibility of
the idea. We compared particle-based gradients with our new estimators
to PILCO. In our previous work \cite{pipps}, we had to change the cost
function to obtain reliable results using particles---one of the
primary motivations of the current experiments was to match PILCO's
results using the same cost as the original PILCO had used (this is
explained in greater detail in Section~\ref{discuss}).

\subsection{Model-based policy search background}

We consider a model-based analogue to the model-free policy search
methods introduced in Section~\ref{reltopol}. The corresponding
probabilistic computation graph is given in
Fig.~\ref{modelbasedpol}. Our notation follows our previous work
\cite{pipps}. After each episode all of the data is used to learn
separate Gaussian process models \cite{gpbook} of each dimension of
the dynamics, s.t.
$\p{\Delta x_{t+1}^a} = \mathcal{GP}(\tilde{\bfv{x}}_t)$, where
$\tilde{\bfv{x}} = [\bfv{x}_t^T,\bfv{u}_t^T]^T$ and
$\bfv{x}\in\mathbb{R}^D$, $\bfv{u}\in\mathbb{R}^F$.  This model is
then used to perform "mental simulations" between the episodes to
optimise the policy by gradient descent. We used a squared exponential
covariance function
$k_a(\tilde{\bfv{x}},\tilde{\bfv{x}}') = s_a^2\exp(-(\tilde{\bfv{x}} -
\tilde{\bfv{x}}')^T\Lambda_a^{-1}(\tilde{\bfv{x}} -
\tilde{\bfv{x}}'))$. We use a Gaussian likelihood function, with noise
hyperparameter $\sigma_{n,a}^2$. The hyperparameters,
$\{s, \Lambda, \sigma_n\}$ are trained by maximizing the marginal
likelihood. The predictions have the form
$\textup{p}(\bfv{x}_{t+1}^a) = \mathcal{N}(\mu(\tilde{\bfv{x}}_t),
\sigma_f^2(\tilde{\bfv{x}}_t)+\sigma_n^2)$, where
$\sigma_f^2(\tilde{\bfv{x}}_t)$ is an uncertainty about the model, and
depends on the availability of data in a region of the state-space.

\subsection{Setup}

The cart-pole consists of a cart that can be pushed back and
forth, and an attached pole. The state space is
$[s, \beta, \dot{s}, \dot{\beta}]$, where $s$ is the cart position and
$\beta$ the angle. The control is a force on the cart. The dynamics
were the same as in a PILCO paper \cite{deisenroth2015gppilco}. The
setup follows our prior work \cite{pipps}.

\paragraph{Common properties in tasks} The experiments consisted of 1
random episode followed by 15 episodes with a learned policy, where
the policy is optimized between episodes. Each episode length was 3s,
with a 10Hz control frequency. Each task was evaluated separately 100
times with different random number seeds to test repeatability. The
random number seeds were shared across different algorithms. Each
episode was evaluated 30 times, and the cost was averaged, but note
that this was done only for evaluation purposes---the algorithms only
had access to 1 episode. The policy was optimized using an
RMSprop-like learning rule \cite{tieleman2012rmsprop} from our
previous work \cite{pipps}, which normalizes the gradients using the
sample variance of the gradients from different particles. In the
model-based policy optimization, we performed 600 gradient steps using
300 particles for each policy gradient evaluation. The learning rate
and momentum parameters were $\alpha = 5\times10^{-4}$, $\gamma = 0.9$
respectively---the same as in our previous work. The output from the
policy was saturated by $\textup{sat}(u) = 9\sin(u)/8 + \sin(3u)/8$, where
$u = \tilde{\pi}(\bfv{x})$. The policy $\tilde{\pi}$ was a radial
basis function network (a sum of Gaussians) with 50 basis functions
and a total of 254 parameters. The cost functions were of the type
$1 - \exp(-(\bfv{x}-\bfv{t})^TQ(\bfv{x}-\bfv{t}))$, where $\bfv{t}$ is
the target. We considered two types of cost functions: 1)
\textit{Angle Cost}, a cost where $Q = \textup{diag}([1,1,0,0])$ is a
diagonal matrix, 2) \textit{Tip Cost}, a cost from the original PILCO
papers, which depends on the distance of the tip of the pendulum to
the position of the tip when it is balanced. These cost functions are
conceptually different---with the \textit{Tip Cost} the pendulum
could be swung up from either direction, with the \textit{Angle Cost}
there is only one correct direction.  The base observation noise
levels were $\sigma_s = 0.01 ~\textup{m}$,
$\sigma_\beta = 1 ~\textup{deg}$,
$\sigma_{\dot{s}} = 0.1 ~\textup{m}/\textup{s}$,
$\sigma_{\dot{\beta}} = 10 ~\textup{deg}/\textup{s}$, and these were
modified with a multiplier $k\in\{10^{-2},1\}$, such that
$\sigma^2=k\sigma^2_{base}$.

\paragraph{Cart-pole swing-up and balancing} In this task the pendulum
starts hanging downwards, and must be swung up and balanced. We
took some results from our previous work \cite{pipps}: PILCO;
reparameterization gradients (RP); Gaussian resampling (GR);
batch importance weighted LR, with a batch importance weighted
baseline (LR); total propagation combining BIW-LR and RP (TP). We
compared to the new methods: Gaussian shaping gradients using
the BIW-LR component (GLR), Gaussian shaping gradients combining
BIW-LR and RP variants using total propagation (GTP). Moreover, we tested GTP
when the model noise variance was multiplied by 25 (GTP$+\sigma_n$).

\paragraph{Cart-pole balancing with DEL estimator} This task is much
simpler---the pole starts upright and must be balanced.
The experiment was devised to show that DEL is feasible
and may be useful if further developed. The \textit{Angle Cost}
and the base noise level were used.

\subsection{Results}

\begin{figure}[!t]
        \centering
	\begin{subfigure}{.32\textwidth}
		\includegraphics[width=\textwidth]{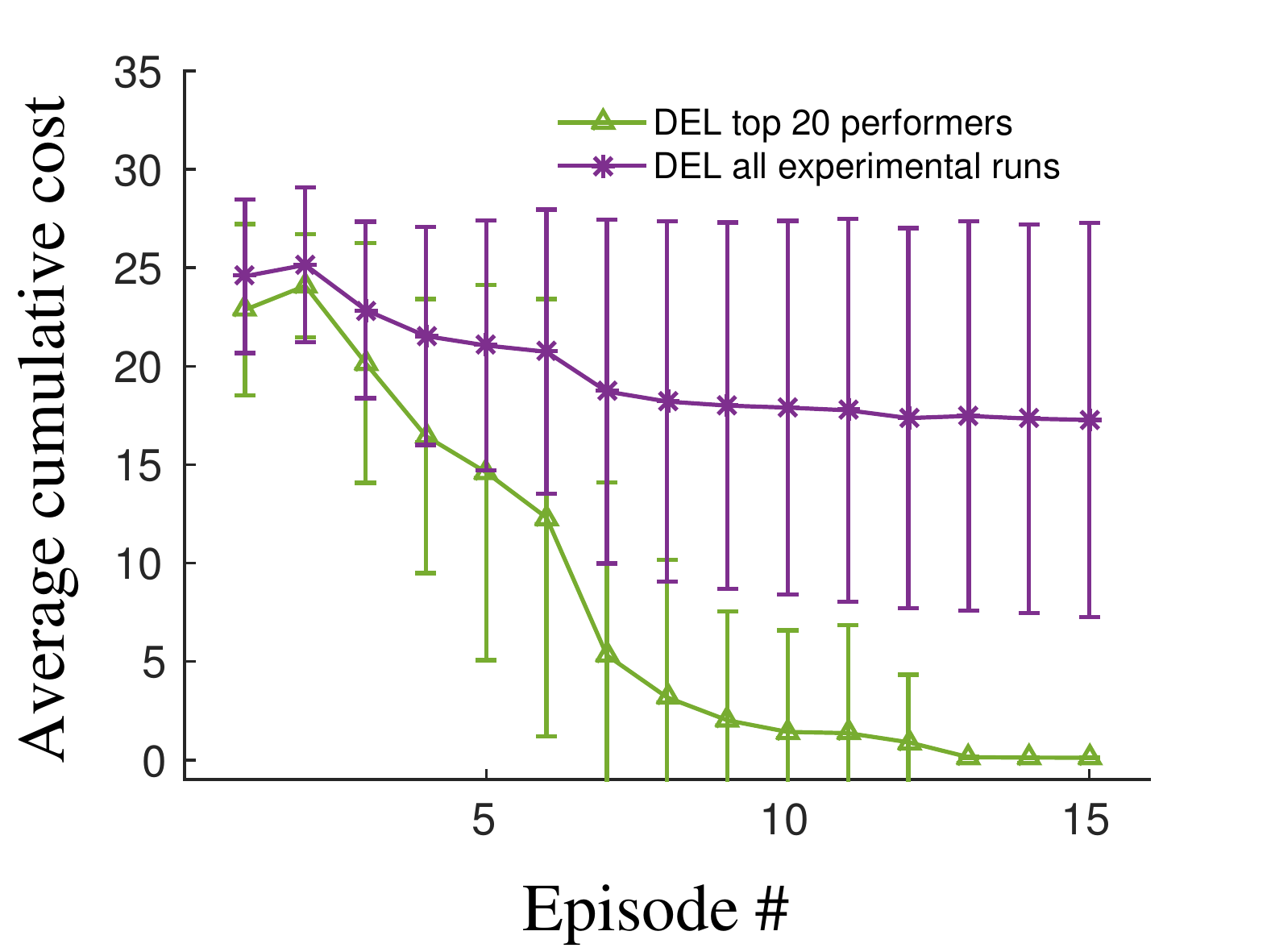}
		\caption{Cart-pole balancing only\\\mbox{}}
          \label{delbalancing}
        \end{subfigure}
	\begin{subfigure}{.32\textwidth}
		\includegraphics[width=\textwidth]{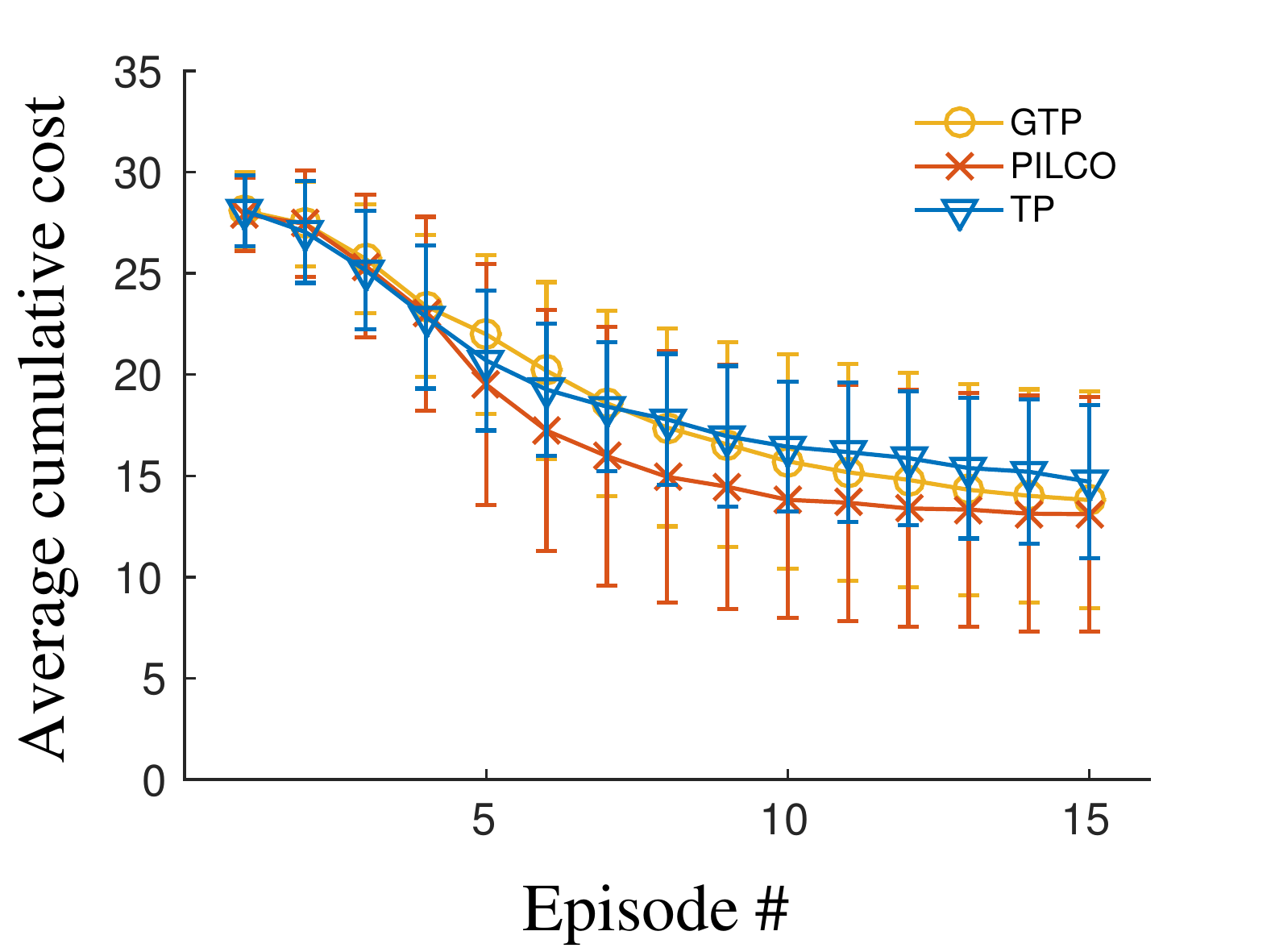}
		\caption{Swing-up and balancing\\ \mbox{\hspace{2.4ex}}
                  All experimental runs}
          \label{swingupall}
	\end{subfigure}
	\begin{subfigure}{.32\textwidth}
		\includegraphics[width=\textwidth]{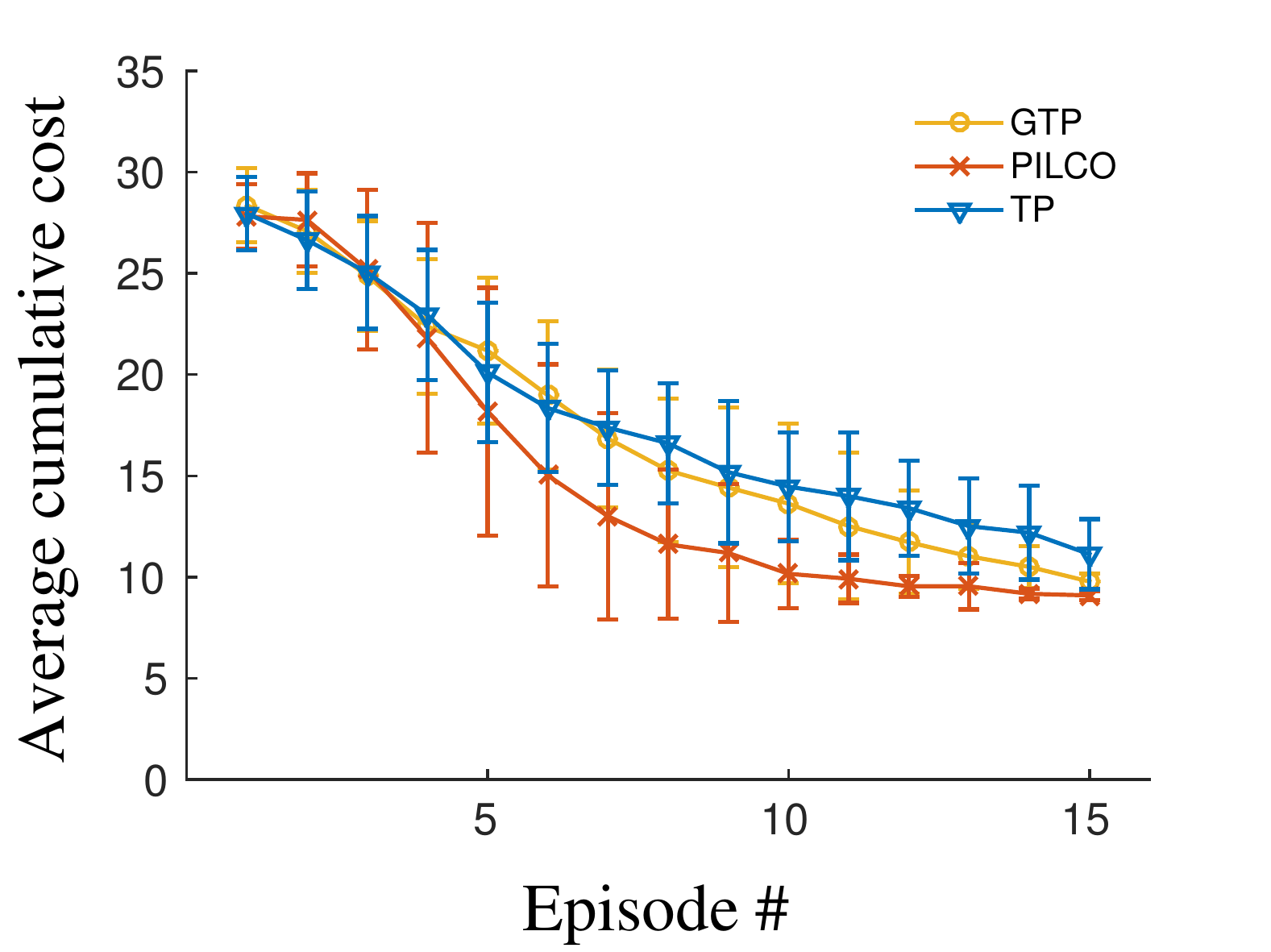}
		\caption{Swing-up and balancing\\ \mbox{\hspace{2.4ex}}
                  Top 40 experimental runs}
          \label{swinguppeak}
        \end{subfigure}
	\caption{Data-efficiency and performance of learning
          algorithms on cart-pole tasks. Figures~\ref{swingupall} and
          \ref{swinguppeak} correspond to the $k=1$, \textit{Tip Cost}
          case.}
          \label{learningeff}
\end{figure}

The results are presented in Table~\ref{cpexps} and in
Fig.~\ref{learningeff}. Similarly to our previous work \cite{pipps},
with low noise, methods which include LR components do not
work well. However, the GTP$+\sigma_n$ experiments show that injecting
more noise into the model predictions can solve the problem. The main
important result is that GTP matches PILCO in the \textit{Tip Cost}
scenarios. In our previous work \cite{pipps}, one of the concerns was
that TP had not matched PILCO in this scenario. Looking only at the
costs in Fig.~\ref{swingupall} and \ref{swinguppeak} does not
adequately display the difference. In contrast, the success rates show
that TP did not perform as well. The success rates were measured both
by a threshold which was calibrated in previous work (final loss below
15) as well as by visually classifying all experimental runs. Both
methods agreed. The losses of the peak performers at the final episode
were ~~~TP: $11.14\pm1.73$,~~~ GTP: $9.78\pm0.40$, ~~~PILCO:
$9.10\pm0.22$, which also show that TP was significantly worse. While
the peak performers were still improving, the remaining experiments
had converged. PILCO still appears slightly more data-efficient;
however, the difference has little practical significance as the
required amount of data is low. Also note that in
Fig.~\ref{swingupall} TP has smaller variance. The larger variance
of GTP and PILCO is caused by outliers with a large loss. These
outliers converged to a local minimum, which takes advantage of the
tail of the Gaussian approximation of the state distribution---this
contrasts with prior suggestions that PILCO performs exploration using
the tail of the Gaussian \cite{deisenroth2011pilco}.

\begin{table}
  \caption{Success rate of learning cart-pole swing-up}
\vskip 0.1in
\label{cpexps}
\centering
\begin{tabular}{llllllllll}
\toprule
  Cost func. & $\sigma_o^2$ multiplier & PILCO & RP & GR & LR & TP & GTP & GLR & GTP$+\sigma_n$\\
\midrule
Angle Cost & $k = 10^{-2}$ & {\bf 0.88} & 0.69 & 0.63 & 0.57 & {\bf 0.82} & 0.65 & 0.42 & {\bf 0.88}\\
  Angle Cost & $k = 1$ & 0.79 & 0.74 & 0.89 & {\bf 0.96} & {\bf 0.99} & {\bf 0.9} & {\bf 0.93} &\\
  Tip Cost & $k = 10^{-2}$ & {\bf 0.92} & 0.44 & 0.47 & 0.36 & 0.54 & 0.6 & 0.45 & 0.8\\
Tip Cost & $k = 1$ & {\bf 0.73} & 0.15 & {\bf 0.68} & 0.28 & 0.48 & {\bf 0.69} & 0.35 &\\
\bottomrule
\end{tabular}
\vskip -0.1in
\end{table}

\subsection{Discussion}
\label{discuss}
Our work demystifies the factors which contributed to the success of
PILCO.  It was previously suggested that the Gaussian approximations
in PILCO smooth the reward, and cause unimodal trajectory
distributions, simplifying the optimization problem
\cite{andrew,gal2016nnpilco}. In our previous work \cite{pipps},
we showed that the
main advantage was actually that it prevents the curse of
chaos/exploding gradients. In the current work we
decoupled the gradient and reward effects, and provided evidence that
both factors contributed to the success of Gaussian distributions. While GR
often has similar performance to GTP, there is an important conceptual
difference: GR performs resampling, hence the trajectory distribution
is not an estimate of the true trajectory distribution. Moreover,
unlike resampling, GTP does not remove the temporal dependence in
particles, which may be important in some applications.

\section{Conclusions \& future work}

We have created an intuitive graphical framework for visualizing and
deriving gradient estimators in a graph of probabilistic
computations. Our method provides new insights towards previous policy
gradient theorems in the literature. We derived new
gradient estimators based on density estimation (DEL), as well as
based on the idea to perform a \emph{jump estimation} to an
intermediate node, not directly to the expected cost (GS). The DEL
estimator needs to be
further developed, but it has good conceptual properties
as it should not suffer from the curse of chaos nor does it require
injecting noise into computations. The GS estimator allows
differentiating through discrete computations in a manner that will
still allow backpropagating pathwise derivatives. Finally, we provided
additional evidence towards demystifying the success of the
popular PILCO algorithm. We hope that our work could lead towards new
automatic gradient estimation software frameworks which are not only
concerned with computational speed, but also the accuracy of the estimated
gradients.

\subsubsection*{Acknowledgments}

We thank the anonymous reviewers for useful comments.
This work was supported by OIST Graduate School funding and by
JSPS KAKENHI Grant Number
JP16H06563 and JP16K21738.

\bibliography{totalgrad}
\bibliographystyle{apalike}

\end{document}